\documentclass[11pt,a4paper]{article}
\usepackage[hyperref]{acl2020}
\usepackage{times}
\usepackage{latexsym}

\usepackage{microtype}

\aclfinalcopy 

\title{Demo of the Linguistic Field Data Management and Analysis System - LiFE}

\author{Siddharth Singh, Ritesh Kumar, Shyam Ratan, Sonal Sinha \\
  Department of Linguistics, Dr. Bhimrao Ambedkar University,   Agra \\
  \texttt{sidd435@gmail.com} \quad
  \texttt{ritesh78\_llh@jnu.ac.in} \quad \\
  \texttt{shyamratan2907@gmail.com} \quad
  \texttt{sonalsinha2612@gmail.com} \\}

\date{}

\begin{document}
\maketitle
\begin{abstract}
In the proposed demo, we will present a new software - Linguistic Field Data Management and Analysis System - LiFE \footnote{\url{https://github.com/kmi-linguistics/life}} - an open-source, web-based linguistic data management and analysis application that allows for systematic storage, management, sharing and usage of linguistic data collected from the field. The application allows users to store lexical items, sentences, paragraphs, audio-visual content including photographs, video clips, speech recordings, etc, along with rich glossing / annotation; generate interactive and print dictionaries; and also train and use natural language processing tools and models for various purposes using this data. Since its a web-based application, it also allows for seamless collaboration among multiple persons and sharing the data, models, etc with each other. 

The system uses the Python-based Flask framework and MongoDB (as database) in the backend and HTML, CSS and Javascript at the frontend. The interface allows creation of multiple projects that could be shared with the other users. At the backend, the application stores the data in RDF format so as to allow its release as Linked Data over the web using semantic web technologies - as of now it makes use of the OntoLex-Lemon for storing the lexical data and Ligt for storing the interlinear glossed text and then internally linking it to the other linked lexicons and databases such as DBpedia and WordNet. Furthermore it provides support for training the NLP systems using scikit-learn and HuggingFace Transformers libraries as well as make use of any model trained using these libraries - while the user interface itself provides limited options for tuning the system, an externally-trained model could be easily incorporated within the application; similarly the dataset itself could be easily exported into a standard machine-readable format like JSON or CSV that could be consumed by other programs and pipelines. The system is built as an online platform; however since we are making the source code available, it could be installed by users on their internal / personal servers as well.
\end{abstract}



\section{Introduction}

Linguistic data management and analysis tools have always been a requirement of field linguists. A huge amount of data is collected and analysed by field linguists for a large number of languages including relatively lesser-known, minoritised and endangered languages of the world and these need to be properly stored, analysed and made accessible to the larger community. On the other hand, there are a huge number of languages across the globe (including the kinds mentioned above), whose data is not available for building any kind of language technology tools and applications.
In order to tackle this multi-faceted problem of storing, processing, retrieving and analysing the primary linguistic data, an integrated system with an easily-accessible and user-friendly interface aimed at linguists needs to be made available. ``\verb|LiFE|'' is developed with the intent of providing a practical intervention in the field by making available an organised framework for management, analysis, sharing (as linked data) and processing of primary linguistic field data including development of digital and print lexicons, sketch grammars and fundamental language processing tools such as part-of-speech tagger and morphological analysers. The software provides an easy-to-use, intuitive interface for performing all the tasks and there is an emphasis on automating the tasks as far as possible. For example, given some initial input, the system incrementally trains automated methods for inter-linear glossing of the dataset (which improves as more data is stored in the system) and subsequent generation of sketch grammar as well as NLP tools for the language. Similarly, the system automatically infers and links the entries in the lexicon and inter-linear glossed data using Lemon (more specifically OntoLex-Lemon) \cite{McCrae17} and Ligt \cite{Chiarcos19}.


\section{Motivation and Features}

Linguistic field data storage, management, sharing and linked data generation has largely developed independent of each other. As such while there are quite a few tools and applications aimed at field linguists (or community members interested in fieldwork for their own language) for collection and management of data as well as generating lexicon, such as FieldWorks Language Explorer (FLEx)\footnote{https://software.sil.org/fieldworks/} 
\cite{Butler07} \cite{Manson20}; Toolbox\footnote{https://software.sil.org/shoebox/, https://software.sil.org/toolbox/} 
\cite{Robinson07}; LexiquePro\footnote{https://software.sil.org/lexiquepro/} 
\cite{Guerin07}; WeSay\footnote{https://software.sil.org/wesay/} 
\cite{Perlin12} \cite{Albright08} and a few other platforms for archiving and providing access to the data, the prominent ones being Endangered Languages Archive (ELAR)\footnote{https://www.elararchive.org/} 
\cite{Nathan10}; The Language Archive (TLA)\footnote{https://archive.mpi.nl/tla/} 
\cite{cho2012}; SIL Language and Culture Archive\footnote{https://www.sil.org/resources/language-culture-archives}, etc. The Open Language Archives Community (OLAC)\footnote{http://www.language-archives.org/archives}, which is a consortium of over 60 participating linguistic archives of various kinds (including the ones mentioned above and others for storage and access of linguistic data, especially of endangered languages) has also recently joined the Linguistic Linked Data Open Cloud which paves the way for providing a large amount of such data as linked data \cite{Simons03}. However none of the tools and platforms directly provide an interface for storing or (largely) automatically generating the primary linguistic data as linked data or provide a seamless two-way between the NLP tools and libraries and linguistic data management softwares.

On the other hand, the linked data community has developed tools for supporting generation of linked data, especially linked data lexicons. One of the best-known tools for this is
\textbf{VocBench (VB)}, which is a fully-fledged open-source web-based thesaurus management platform with the feature of collaborative development of multilingual datasets compatible with semantic Web standards. It provides the facilities of generating lexicons, thesauri, and linked data ontologies to the large organisations, companies, and user communities \cite{Stellato20}. However tools like these focus on generating Linked Data which is generally not very user-friendly for field linguists nor do they provide options for automating the tasks or linking to the NLP ecosystem.
The primary motivation for building this platform is to provide a tool that acts as a bridge between field linguists (who are primarily engaged in data collection from low-resource and endangered languages, building lexicons, writing grammatical descriptions and also producing educational and other kinds of materials for the communities that they work with), linked data community (who are primarily engaged in meaningfully connecting data from different languages and resources using the semantic web techniques) and the NLP community (who primarily makes use of the linguistic data from multiple languages; could potentially provide support in automating the tasks carried out by field linguists; and also provide tools and technologies for the marginalised and under-privileged linguistic communities). As such in its current state the app provides the following functionalities -

\begin{itemize}
    \item It provides a  user-friendly interface for storing, sharing and making publicly available the linguistic field data including interlinear glossed text, lexicon and associated multimedia content.
    
    \item It provides reasonable automation for tasks such as generating lexicon, sketch grammar, etc by providing interfaces for training as well as using pre-trained NLP models needed for automating various tasks. The tool currently supports training various algorithms of the scikit-learn and HuggingFace Transformers library as well as using the models trained using these libraries.
    
    \item It provides interface for exporting the data in structured formats such as RDF, JSON and CSV that could be directly used for NLP experiments and modelling.
\end{itemize}

During the demo we will present these features and the interface of the tool in detail and also briefly train the participants in using it.

\section{Presenters}

The demo will be given by the developers of this application which include the following -

\begin{enumerate}
    \item \textbf{Ritesh Kumar} is Assistant Professor of Linguistics and coordinator of the masters program in computational linguistics at Dr. Bhimrao Ambedkar University, Agra. he is working in the field of computational linguistics and language documentation and description for over last 10 years. He has conceptualised, mentored and co-developed this app.
    \item \textbf{Siddharth Singh} is a software engineer and is currently pursuing his MSc in Computational Linguistics from Dr. Bhimrao Ambedkar University. He is the principal developer of the app,
    \item \textbf{Shyam Ratan} is pursuing his Mphil in Computational Linguistics and is a co-developer of the app.
    \item \textbf{Sonal Sinha}  is pursuing her Mphil in Computational Linguistics and is a co-developer of the app.
\end{enumerate}
 
\bibliography{anthology,acl2020}
\bibliographystyle{acl_natbib}

\end{document}